\documentclass[lettersize,journal]{IEEEtran}
\usepackage{amsmath,amsfonts}
\usepackage{array}
\usepackage[caption=false,font=normalsize,labelfont=sf,textfont=sf]{subfig}
\usepackage{textcomp}
\usepackage{stfloats}
\usepackage{url}
\usepackage{verbatim}
\usepackage{graphicx}
\def\BibTeX{{\rm B\kern-.05em{\sc i\kern-.025em b}\kern-.08em
    T\kern-.1667em\lower.7ex\hbox{E}\kern-.125emX}}
\usepackage{balance}

\usepackage{array}
\usepackage{booktabs} 
\usepackage{multirow}
\usepackage{makecell}
\usepackage{soul}
\usepackage{soulpos}
\usepackage{subcaption}
\usepackage{algorithm}
\usepackage{algpseudocode}
\usepackage{amsmath}
\usepackage{algorithm}
\usepackage{float}
\usepackage[algo2e]{algorithm2e} 
\usepackage{xcolor}
\usepackage{cite}

\definecolor{bluecolor}{rgb}{0.21,0.49,0.74}
\usepackage[pagebackref,breaklinks,colorlinks,allcolors=bluecolor]{hyperref}
\usepackage{cleveref}

\begin{document}
\title{Towards Adaptive Human-centric Video Anomaly Detection: A Comprehensive Framework and A New Benchmark}
\author{
 Armin Danesh Pazho$^*$$^\dagger$$^{1}$, Shanle Yao$^*$$^{1}$, Ghazal Alinezhad Noghre$^*$$^{1}$, Babak Rahimi Ardabili$^{2}$, Vinit Katariya$^{1}$, Hamed Tabkhi$^{1}$
\thanks{$^*$ Equal contribution.}
\thanks{$^\dagger$ Corresponding author (adaneshp@charlotte.edu).}
\thanks{$^{1}$ Department of Electrical and Computer Engineering, University of North Carolina at Charlotte ([syao, galinezh, vkatariy, htabkhiv]@charlotte.edu}
\thanks{$^{2}$ Public Policy Program, University of North Carolina at Charlotte (brahimia@charlotte.edu)}
}

\markboth{HuVAD}%
{Towards Adaptive Human-centric Video Anomaly Detection: A Comprehensive Framework and A New Benchmark}

\maketitle

\begin{abstract}
Human-centric Video Anomaly Detection (VAD) aims to identify human behaviors that deviate from normal. At its core, human-centric VAD faces substantial challenges, such as the complexity of diverse human behaviors, the rarity of anomalies, and ethical constraints. These challenges limit access to high-quality datasets and highlight the need for a dataset and framework supporting continual learning. Moving towards adaptive human-centric VAD, we introduce the HuVAD (Human-centric privacy-enhanced Video Anomaly Detection) dataset and a novel Unsupervised Continual Anomaly Learning (UCAL) framework. UCAL enables incremental learning, allowing models to adapt over time, bridging traditional training and real-world deployment. HuVAD prioritizes privacy by providing de-identified annotations and includes seven indoor/outdoor scenes, offering over $5\times$ more pose-annotated frames than previous datasets. Our standard and continual benchmarks, utilize a comprehensive set of metrics, demonstrating that UCAL-enhanced models achieve superior performance in $82.14\%$ of cases, setting a new state-of-the-art (SOTA). The dataset can be accessed at \href{https://github.com/TeCSAR-UNCC/HuVAD}{https://github.com/TeCSAR-UNCC/HuVAD}.
\end{abstract}


\section{Introduction}
\label{sec:intro}

Human-centric Video Anomaly Detection (VAD) refers to identifying events or patterns in human behavior that deviate from the expected behavior. Advancing human-centric VAD technologies encounter several substantial challenges emerging from its context-specific and open-set nature, where anomalies vary across different environments and new unseen events regularly happen, making it difficult to generalize VAD algorithms \cite{noghre2023understanding, noghre2024exploratory, pazho2023survey}. Furthermore, anomalous events are, by definition, rare, creating a scarcity of positive samples in datasets, making it challenging for models to learn solely from labeled examples, pushing the field towards unsupervised approaches \cite{doshi2023towards, danesh2023chad}.

\begin{figure*}
    \centering
       \includegraphics[width=\textwidth]{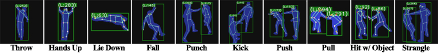}
            \caption{Sample of anomalies and their annotations in the new proposed benchmark: HuVAD dataset. Cropped for visualization purposes. Segmentation is solely used for demonstration purposes.}
            
 \label{fig:intro}
\end{figure*}

To adjust models to tackle such challenges, and adapt to real-world scenarios, research is moving towards continual learning \cite{doshi2020continual, doshi2021online, doshi2022rethinking, doshi2023towards, zhu2024rcl}. Especially, context-specificity and open-set nature of human-centric VAD underscore a critical need for a new framework and dataset to address the data limitation and support continual learning \cite{yang2024context, faber2024lifelong}. Such a framework would allow VAD systems to adapt to new patterns over time, thereby improving detection accuracy in dynamic, real-world environments. As a result, addressing these challenges requires a holistic approach to data curation and framework design.

To support extensive data access and facilitate continual learning, we introduce HuVAD, the largest continuously recorded VAD dataset (see samples in \cref{fig:intro}). Effective continual learning requires datasets that mirror real-world surveillance, capturing diverse environments with ongoing recording and ample samples. HuVAD provides over 5x the training frames and 4x the testing frames compared to previous datasets, supporting unsupervised learning and continual model adaptation \cite{Yao_2024_CVPR, saha2023continual, hsu2018re}. It maintains real-world fidelity with continuous recordings of commuters' real-world traffic data across seven scenes, including indoor/outdoor spaces. While 6 cameras capture typical environments within a community space with, a specialized Context-Specific Camera (CSC) observes law enforcement training. This unique setup allows HuVAD to evaluate model performance in scenarios where normal behaviors differ from typical public environments. To address potential privacy concerns and biases against minority groups \cite{saheb2023ethically, ardabili2023understanding, joshi2024synthetic, kunchala2023towards}, HuVAD employs anonymization, publishing only de-identified human annotations, such as bounding boxes, tracking IDs, and poses \cite{Hirschorn_2023_ICCV, noghre2024exploratory, noghre2023understanding}.


With HuVAD providing an ideal environment for continual learning, we introduce a novel Unsupervised Continual Anomaly Learning (UCAL) framework. UCAL tackles VAD's context-specificity and open-set nature by incrementally adapting the model for each environment in the HuVAD dataset, allowing it to capture and evolve with dynamic patterns. The framework simulates real-world streaming data by organizing data per camera and randomly injecting anomalies into the training stream. Structured in multiple incremental training steps, UCAL leverages a pre-trained model to refine its performance progressively, with continual evaluation against the HuVAD test set.

For a thorough benchmark, we statistically compare HuVAD with peer datasets and evaluate state-of-the-art (SOTA) pose-based VAD algorithms. Unlike prior works that rely on a single metric \cite{yu2023regularity, huang2022hierarchical, zeng2021hierarchical, chen2023multiscale, jain2021posecvae, markovitz2020graph, Hirschorn_2023_ICCV}, we evaluate these algorithms using a comprehensive set of metrics, including Area Under the Receiver Operating Characteristic Curve (AUC-ROC), Area Under the Precision-Recall Curve (AUC-PR), Equal Error Rate (EER), and the 10\% Error Rate (10ER), a metric brought to human-centric VAD for the first time to reflect the unequal costs of false positives and negatives in real-world scenarios. Recognizing the importance of continual learning, we conduct a tailored benchmark of the UCAL framework on the HuVAD dataset. Results demonstrate that models enhanced by the UCAL framework achieve superior performance in 82.14\% of cases, setting a new SOTA in human-centric VAD.

The contributions of this paper are:
\begin{itemize}

    \item HuVAD, the largest continuously recorded dataset in real-world community spaces as well as a especial context-specific law enforcement training environment, emphasizing the impact of context awareness and social interaction on VAD, with comprehensive de-identified human annotations to ensure privacy.
    \item UCAL, a novel Unsupervised Continual Anomaly Learning framework, enabling adaptive learning for VAD by allowing models to evolve continuously with new data.
    \item Providing detailed statistical comparisons, conventional algorithmic evaluations, and offering the first comprehensive benchmark for continual learning in human-centric VAD, showcasing SOTA results of UCAL-enhanced models in $82.14\%$ of cases.
\end{itemize}

\section{Related Works}
\label{sec:related}



\subsection{Video Anomaly Detection Dataset}

In this paper, we focus on unsupervised VAD, which differs fundamentally from weakly-supervised tasks \cite{Cao_2023_CVPR} addressed by datasets such as UCF-Crime \cite{sultani2018real} and XD-Violence \cite{wu2020not}. Additionally, UBnormal \cite{acsintoae2022ubnormal}, composed entirely of synthetic videos, may not reliably reflect real-world environments, limiting its relevance for unsupervised detection. Datasets such as UCF-Crime, XD-Violence, MSAD \cite{zhu2024advancingvideoanomalydetection}, and CUVA \cite{Du_2024_CVPR}, compiled from varied sources rather than continuous recordings, reflect a different task formulation and are thus not comparable to our approach \cite{ramachandra2020street}.

A few recent datasets feature anomalies for both vehicles and pedestrians. Street Scene \cite{ramachandra2020street} dataset is distinguished by its near bird's eye view perspective of a street, and includes non-human anomalies such as illegally parked cars. The NOLA dataset \cite{doshi2022rethinking} with over 1.4 million frames is captured from a single moving camera. In addition to human-centric anomalies, it includes anomalies such as a vehicle moving in the wrong direction.

Several early datasets started the field of VAD such as the Subway dataset \cite{Subway}, the UCSD Pedestrian dataset \cite{UCSD}, and the CUHK Avenue dataset \cite{lu2013abnormal}. While pivotal for VAD's early development, these datasets are now considered relatively small and feature a limited variety of scenes.


More recent datasets have emerged to advance the field of VAD. The ShanghaiTech Campus (SHT) dataset \cite{liu2018future} stands as a primary benchmark within the realm of VAD, particularly for human-centric approaches. It includes anomalies such as chases and fights. A shortcoming of the dataset is that the frames per camera are limited, posing challenges for continual learning. The IITB dataset \cite{rodrigues2020multi} captures human activities within a corridor, recorded using a single fish-eye camera. The CHAD dataset \cite{danesh2023chad} is a large-scale VAD dataset recorded within a parking lot setting. CHAD features 22 anomaly classes and is captured using four high-resolution cameras at 30 FPS. The NWPUC dataset \cite{Cao_2023_CVPR} was developed to introduce greater diversity in a campus setting, featuring 28 classes of anomalies across 43 distinct scenes. However, the dataset's per-scene volume remains relatively limited, which may constrain its comprehensiveness for certain large-scale anomaly detection tasks such as continual learning.




\subsection{Continual Learning}


Continual learning has been applied to anomaly detection across various fields, addressing challenges such as data drift and evolving data streams. Bugarin et al. \cite{bugarin2024unveilingct} introduce a benchmark for anomaly detection in industrial images, while Chavan et al. \cite{chavan2024activect} propose a framework to handle data drift in industrial settings. Mozaffari et al. \cite{mozaffari2023selfct} develop an approach for high-dimensional data streams in cybersecurity, and Mazarbhuiya and Shenify \cite{mazarbhuiya2023realct} focus on clustering for real-time anomaly detection in IoT. Together, these works showcase continual learning’s potential to enhance anomaly detection in complex, evolving environments.

\begin{figure*}
    \centering
       \includegraphics[width=1\linewidth, trim= 0 0 0 0,clip]{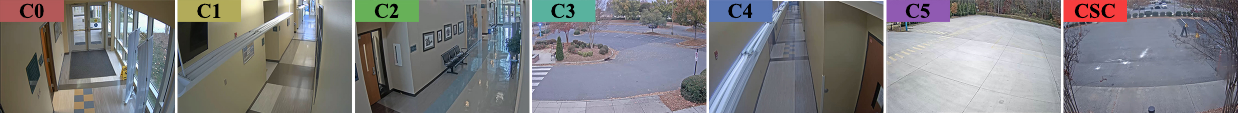}
            \caption{The camera views excluding people. The ratio has been adjusted to fit the manuscript.}
            
 \label{fig:camview}
\end{figure*}

\begin{figure}
    \centering
       \includegraphics[width=0.81\columnwidth, trim= 0 0 0 0,clip]{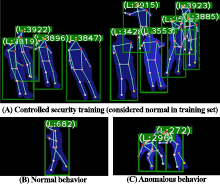}
            \caption{Samples from CSC Camera. Segmentation is solely used for demonstration purposes.}
            
 \label{fig:csc}
\end{figure}

Initial efforts to integrate continual learning into VAD are led by Doshi and Yilmaz, who developed several frameworks enhancing VAD through continual learning. Their 2020 work \cite{doshi2020continual} combines transfer learning with k-nearest neighbor (kNN) techniques for low-complexity, continual anomaly detection. In 2022 \cite{doshi2022rethinking} they focus on pedestrian and vehicle VAD with continual and few-shot learning to improve detection delay and alarm precision. Their 2021 MONAD framework \cite{doshi2021online} integrates GAN-based prediction with statistical guarantees on false alarms, ensuring robustness. Although these works contribute significantly to adaptive VAD, progress can accelerate further with the establishment of a unified and comprehensive benchmark, particularly in continual human-centric VAD.

\section{Privacy and Ethical Considerations} \label{sec:privacy}

Integrating computer vision algorithms into various sectors of society has underscored the importance of responsibly developing these technologies \cite{ahmad2022developing}. VAD and Smart Video Surveillance (SVS) touch upon critical ethical concerns such as bias, discrimination, and privacy violations \cite{ardabili2022understanding}. These concerns have practical implications for the performance and fairness of computer vision systems in the real world. Bias in VAD algorithms can lead to discriminatory practices, where specific demographics may be unfairly targeted or misrepresented \cite{raso2018artificial, noriega2020application}. Privacy concerns are equally significant, as the pervasive monitoring and analysis of individuals without safeguards can lead to unwarranted privacy invasions, raising ethical and legal issues \cite{ardabili2023understanding}.

Addressing these privacy and ethical concerns necessitates a multifaceted approach. Such considerations are particularly crucial during the dataset creation \cite{whang2023data} and algorithm development \cite{ardabili2023understanding}. Adopting more abstract representation techniques, such as only focusing on human pose information, has been proposed as an intermediate solution to address some privacy, biases, and ethical concerns if not fully mitigating them \cite{ardabili2023understanding}. Our interaction with communities also proves that the public and stakeholders are more responsive and participatory to dataset collection when focusing solely on human pose information rather than actual pixels. However, transitioning to a more abstract approach (e.g. human pose rather than pixel-level data) raises questions regarding the potential compromise in model accuracy. Recent advancements challenge this notion. On the SHT dataset \cite{liu2018future}, pixel-based methods like SSMTL++v2 \cite{barbalau2023ssmtl++} and Jigsaw-VAD \cite{wang2022video} achieve AUC-ROC scores of 83.80 and 84.30, while pose-based approaches like MoPRL \cite{yu2023regularity} and STG-NF \cite{Hirschorn_2023_ICCV} attain AUC-ROC scores of 83.35 and 85.90. 

Our objective is to initiate the trend of de-identifying PII at the data collection phase instead of the algorithmic phase. This methodology not only addresses legal challenges but also promotes expedited data gathering and dissemination. Such an approach is anticipated to facilitate research and technological advancements more finely attuned to the complexities of human behavior and societal requirements.
\section{Data Collection and Annotation}
\label{sec:data_collection}

\begin{figure*}
    \centering
       \includegraphics[width=0.7\linewidth, trim= 0 5 0 0,clip]{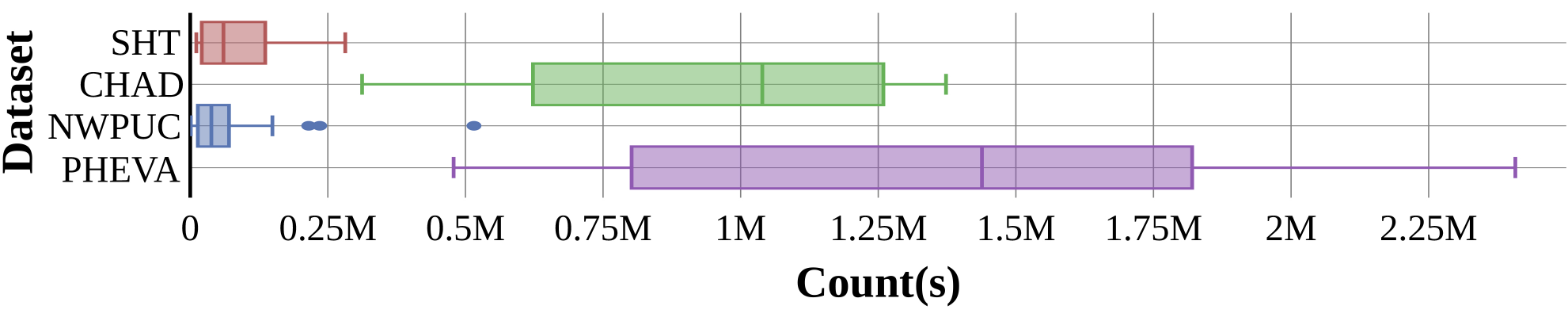}
            \caption{Pose counts per camera across major multi-camera datasets (SHT \cite{liu2018future}, CHAD \cite{danesh2023chad}, and NWPUC \cite{Cao_2023_CVPR}).} 
        \label{fig:bbox_all}
\end{figure*}

\begin{figure*}
    \centering
       \includegraphics[width=0.9\linewidth, trim= 0 0 0 0,clip]{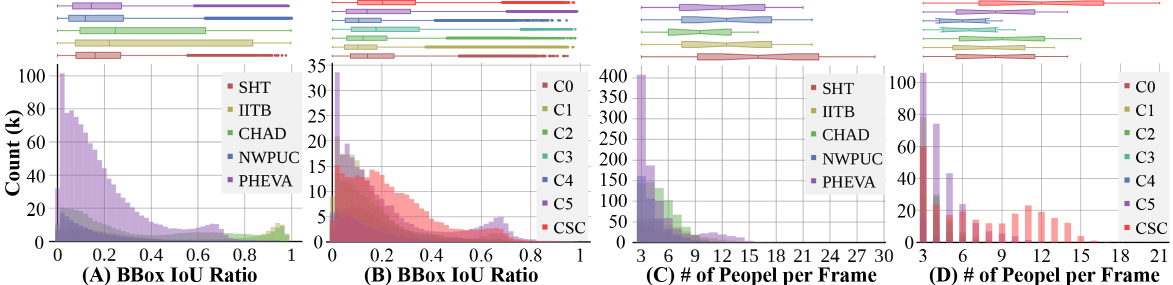}
            \caption{IoU and crowd density across key datasets (SHT \cite{liu2018future}, IITB \cite{rodrigues2020multi}, CHAD \cite{danesh2023chad}, and NWPUC \cite{Cao_2023_CVPR}) and HuVAD's camera views.}
        \label{fig:full_analyze}
\end{figure*} 



The HuVAD dataset was captured using seven Closed-Circuit Television (CCTV) cameras at a resolution of $1280\times720$ pixels, recording over five days from 6:00 AM to 6:00 PM. It includes varied scenes: three outdoor parking areas, three hallways, and a building entrance/exit (see \cref{fig:camview}). Six cameras (C0 to C5) cover normal environments, while a context-specific camera (CSC) focuses on an outdoor area for security and law enforcement training (\cref{fig:csc}). HuVAD includes anomalies such as throwing, hands up, lying down, falling, punching, kicking, pushing, and strangling (see \cref{fig:intro}), enhancing real-world applicability by focusing on behaviors relevant to public safety while excluding less relevant actions such as jumping.


\subsection{Annotation Methodology}
\label{sec:annotation}


\textbf{Anomaly Annotation:} The HuVAD dataset has been meticulously annotated for anomalies to ensure high precision, with each frame reviewed by at least three trained annotators and random testing for label consistency. Anomalous behaviors from \cref{sec:data_collection} are marked abnormal on cameras C0 to C5, while routine activities are labeled as normal. For the CSC camera, if these actions occur during a controlled training session, they are labeled as normal; otherwise, they are labeled as anomalies. The dataset includes frame-level anomaly labels and detailed spatial annotations through region masks that localize anomalies, enabling granular analysis of events. Following the standard approach in unsupervised and self-supervised video anomaly detection benchmarks such as SHT \cite{liu2018future}, IITB \cite{rodrigues2020multi}, CHAD \cite{danesh2023chad}, and NWPUC \cite{Cao_2023_CVPR}, we focus solely on the normal vs. anomalous distinction without class-specific annotations.

\textbf{Peson Annotations:}  For realistic anomaly detection, HuVAD simulates real-world reliance on algorithmically extracted person data. Using a semi-automated approach, human annotations are first generated by models, followed by manual human verification to ensure accuracy.

The HuVAD dataset provides comprehensive de-identified annotations \cite{pazho2023ancilia}, including bounding boxes, person IDs, and pose annotations. Bounding boxes, generated with YOLOv8 \cite{yolov8_ultralytics}, mark detected individuals' positions. Person IDs, extracted with ByteTrack \cite{zhang2022bytetrack}, ensure temporal consistency and identity preservation across frames. Human poses, generated with HRNet \cite{sun2019deep} in the COCO17 format \cite{lin2014microsoft}, capture body keypoints, with linear interpolation and a 15-frame smoothing window for quality assurance, and interpolated poses flagged for optional exclusion.

\label{sec:person_anno}

\subsection{Data Statistics}

HuVAD is the largest continuous VAD dataset to date, with over five million frames ($\sim $46 hours) of annotated pose sequences. As shown in \cref{fig:bbox_all}, HuVAD's per-camera pose counts significantly exceed those of other datasets, providing a robust foundation for developing advanced learning methods crucial for real-world anomaly detection. This extensive data volume per environment directly supports Unsupervised Continual Anomaly Detection (UCAL), discussed in \cref{sec:continual}.

In \cref{fig:full_analyze}.A and \cref{fig:full_analyze}.B, the distribution of the highest Intersection over Union (IoU) ratio per frame highlights HuVAD's occlusion characteristics, offering a challenging environment comparable to NWPUC \cite{Cao_2023_CVPR} and SHT \cite{liu2018future}. Per camera, stats show a high median level of occlusion, particularly in the CSC camera. This diversity in occlusion levels ensures a challenging benchmark for models. \cref{fig:full_analyze}.C and \cref{fig:full_analyze}.D illustrate that HuVAD provides a wide range of crowd densities, from 3 to 15 individuals per frame, with higher densities in CSC due to capturing large-group security exercises. While CHAD \cite{danesh2023chad} and SHT \cite{liu2018future} show variable crowdedness, HuVAD's density distribution is more comprehensive, offering a broad benchmark for crowd-based anomaly detection tasks.

\section{Metrics}

\begin{table*}[]
\centering
\small
\begin{tabular}{@{}cccccccc@{}}
\toprule \toprule
\textbf{Dataset} & \textbf{Conference} & \textbf{Total} & \textbf{Train} & \textbf{Test} & \textbf{Test Normal} & \textbf{Test Anomaly} & \textbf{Camera(s)} \\ \midrule
\multicolumn{1}{c|}{SHT} & CVPR 18 & 295,495 & 257,650 & 37,845 & 21,141 & \multicolumn{1}{c|}{16,704} & 13 \\
\multicolumn{1}{c|}{IITB} & WACV 20 & 459,341 & 279,880 & 179,461 & 71,316 & \multicolumn{1}{c|}{108,145} & 1 \\
\multicolumn{1}{c|}{CHAD} & SCIA 23 & 922,034 & 802,167 & 119,867 & 60,969 & \multicolumn{1}{c|}{58,898} & 4 \\
\multicolumn{1}{c|}{NWPUC} & CVPR 23 & 1,000,129 & 715,901 & 284,228 & 235,957 & \multicolumn{1}{c|}{48,271} & 43 \\ \midrule
\multicolumn{1}{c|}{\textbf{HuVAD (Ours)}} & - & 5,196,675 & 4,467,271 & 729,404 & 517,286 & \multicolumn{1}{c|}{212,118} & 7 \\ \bottomrule \bottomrule 
\end{tabular}%
\caption{Number of frames with at least one human in major datasets (SHT \cite{liu2018future}, IITB \cite{rodrigues2020multi}, CHAD \cite{danesh2023chad}, and NWPUC \cite{Cao_2023_CVPR}). }
\label{tab:frame-count}
\end{table*}

\begin{table*}[]
\centering
\small
\begin{tabular}{@{}c|c|cccc|cccc@{}}
\toprule\toprule
\textbf{Model} & \textbf{Conference} & \textbf{AUC-ROC} & \textbf{AUC-PR} & \textbf{EER} & \textbf{10ER} & \textbf{AUC-ROC} & \textbf{AUC-PR} & \textbf{EER} & \textbf{10ER} \\ \midrule
\multicolumn{2}{c|}{} & \multicolumn{4}{c|}{\textbf{C0}} & \multicolumn{4}{c}{\textbf{C1}} \\ \midrule
\multicolumn{1}{c|}{\textbf{MPED-RNN}} & CVPR 19 & \textbf{79.57} & {\underline{46.76}} & \textbf{0.26} & \textbf{0.37} & 83.57 & {\underline{53.62}} & \textbf{0.22} & \textbf{0.39} \\
\multicolumn{1}{c|}{\textbf{GEPC}} & CVPR 20 & 59.07 & 28.00 & 0.44 & 0.71 & 56.27 & 23.20 & 0.45 & 0.78 \\
\multicolumn{1}{c|}{\textbf{STG-NF}} & ICCV 23 & 58.96 & \textbf{83.45} & 0.46 & 0.84 & 47.82 & \textbf{78.31} & 0.52 & 0.89 \\
\multicolumn{1}{c|}{\textbf{TSGAD}} & WACV 24 & {\underline{64.18}} & 31.93 & {\underline{0.40}} & {\underline{0.66}} & {\underline{68.88}} & 39.02 & {\underline{0.35}} & {\underline{0.72}} \\ \midrule
\multicolumn{2}{c|}{} & \multicolumn{4}{c|}{\textbf{C2}} & \multicolumn{4}{c}{\textbf{C3}} \\ \midrule
\multicolumn{1}{c|}{\textbf{MPED-RNN}} & CVPR 19 & \textbf{73.36} & {\underline{47.66}} & \textbf{0.32} & \textbf{0.57} & 83.62 & {\underline{63.87}} & \textbf{0.23} & \textbf{0.42} \\
\multicolumn{1}{c|}{\textbf{GEPC}} & CVPR 20 & 55.09 & 30.13 & 0.45 & 0.79 & 52.40 & 27.09 & 0.50 & 0.77 \\
\multicolumn{1}{c|}{\textbf{STG-NF}} & ICCV 23 & 51.06 & \textbf{74.20} & 0.49 & 0.94 & 49.15 & \textbf{74.10} & 0.50 & 0.90 \\
\multicolumn{1}{c|}{\textbf{TSGAD}} & WACV 24 & {\underline{62.81}} & 35.53 & {\underline{0.39}} & {\underline{0.73}} & {\underline{54.64}} & 24.25 & {\underline{0.48}} & \underline{0.76} \\ \midrule
\multicolumn{2}{c|}{} & \multicolumn{4}{c|}{\textbf{C4}} & \multicolumn{4}{c}{\textbf{C5}} \\ \midrule
\multicolumn{1}{c|}{\textbf{MPED-RNN}} & CVPR 19 & {\underline{74.59}} & {\underline{44.55}} & \underline{0.29} & \textbf{0.31} & 79.23 & {\underline{49.09}} & \textbf{0.27} & \textbf{0.41} \\
\multicolumn{1}{c|}{\textbf{GEPC}} & CVPR 20 & 72.44 & 30.71 & \underline{0.29} & 0.99 & 66.05 & 33.04 & \underline{0.36} & {\underline{0.56}} \\
\multicolumn{1}{c|}{\textbf{STG-NF}} & ICCV 23 & 72.24 & \textbf{93.95} & \textbf{0.28} & {\underline{0.71}} & 56.48 & \textbf{82.36} & 0.46 & 0.96 \\
\multicolumn{1}{c|}{\textbf{TSGAD}} & WACV 24 & \textbf{75.11} & 37.13 & \textbf{0.28} & 1.00 & {\underline{67.15}} & 35.20 & 0.37 & 0.57 \\ \midrule
\multicolumn{2}{c|}{} & \multicolumn{4}{c|}{\textbf{CSC}} & \multicolumn{4}{c}{\textbf{Combined}} \\ \midrule
\multicolumn{1}{c|}{\textbf{MPED-RNN}} & CVPR 19 & 56.85 & 37.13 & {\underline{0.43}} & \textbf{0.69} & 76.05 & {\underline{42.83}} & \textbf{0.28} & \textbf{0.49} \\
\multicolumn{1}{c|}{\textbf{GEPC}} & CVPR 20 & {\underline{58.32}} & {\underline{41.09}} & \textbf{0.42} & \underline{0.86} & 62.25 & 28.62 & 0.41 & 0.67 \\
\multicolumn{1}{c|}{\textbf{STG-NF}} & ICCV 23 & 53.60 & \textbf{66.28} & 0.47 & 0.92 & 57.57 & \textbf{83.77} & 0.46 & 0.90 \\
\multicolumn{1}{c|}{\textbf{TSGAD}} & WACV 24 & \textbf{58.91} & 43.28 & \underline{0.43} & {\underline{0.86}} & {\underline{68.00}} & 34.61 & {\underline{0.36}} & {\underline{0.64}} \\ \bottomrule \bottomrule
\end{tabular}%
\caption{Benchmark of available SOTA pose-based models (MPED-RNN \cite{morais2019learning}, GEPC \cite{markovitz2020graph}, STG-NF \cite{Hirschorn_2023_ICCV}, and TSGAD \cite{noghre2024exploratory}) on HuVAD-S.}
\label{tab:benchmark}
\end{table*}

AUC-ROC, AUC-PR, EER, and 10ER each provide unique insights and limitations, making their combined use essential for a comprehensive assessment. \textbf{AUC-ROC} measures class distinction ability by plotting the True Positive Rate (TPR) against False Positive Rate (FPR) across thresholds; however, it lacks consideration of the False Negative Rate (FNR), is sensitive to data imbalance, and may obscure key trade-offs \cite{davis2006relationship, fernandez2018learning, he2013imbalanced}. \textbf{AUC-PR} calculates the area under the Precision-Recall curve, which better handles imbalanced datasets but falls short in analyzing negative predictions and overall error distribution \cite{saito2015precision, he2013imbalanced}. \textbf{EER} identifies the threshold where FPR and FNR are equal, offering a balance of sensitivity (detecting anomalies) and specificity (recognizing normals), valuable in real-world deployments where the costs of false positives and negatives are balanced \cite{sultani2018real, li2013anomaly}. Inspired by the False Match Rate 100 (FMR100) metric \cite{maio2002fvc2002, neto2022ocfr}, this paper brings \textbf{10ER} to VAD, measuring the FPR at a fixed 10\% FNR—a threshold regarded as acceptable for VAD \cite{ardabili2024exploring}. 10ER provides a practical, threshold-based perspective and when combined with other metrics, offers a more robust evaluation of VAD performance.

\section{HuVAD Standard Benchmarks (HuVAD-S)}
\label{sec:benchmark}

\begin{figure*}
    \centering
       \includegraphics[width=0.85\linewidth, trim= 20 20 22 20,clip]{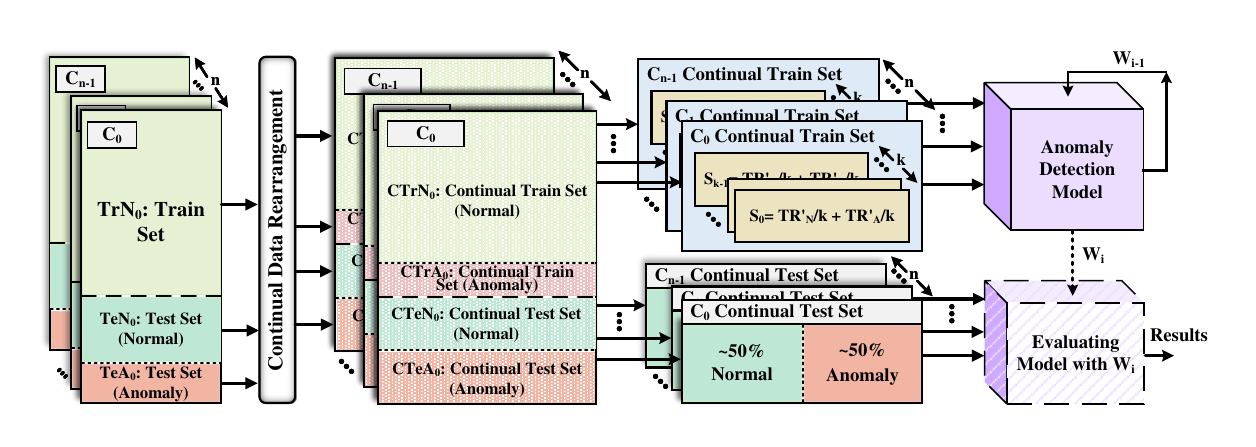}
            \caption{The proposed UCAL framework begins with per-camera data from HuVAD-S, using the Continual Data Rearrangement module to simulate a real-world data stream. The data is then divided into k slices, and the model is incrementally trained on each slice, with performance evaluated on the continual test set at each step. This process is repeated across all cameras.}
        \label{fig:cal}
\end{figure*} 

\begin{algorithm}[]
\small
\caption{UCAL Framework}
\label{algo:continual_learning}

\SetKwInOut{Input}{Input}
\SetKwInOut{Output}{Output}

\Input{$TeA_j$, $TeN_j$, $TrN_j$: Test Anomaly, Test Normal, and Train Normal sets from HuVAD-S for each of $n$ cameras \\
       $k$: Number of data slices per camera \\
       $\theta_{\text{pretrained}}$: Pretrained model weights}
\Output{$CTeA_j$, $CTeN_j$, $CTrN_j$: Updated Test Anomaly, Test Normal, and Train Normal sets for each camera in HuVAD-C}

\For{$j = 1$ \textbf{to} $n$ \textbf{cameras}}{

    \textbf{Step 1: Continual Data Rearrangement} \\
    $CTrA_j \gets \frac{1}{100} \times \mathcal{N}(TrN_j),$\;
    $CTeA_j \gets TeA_j - CTrA_j$\;
    $CTeN_j \gets \mathcal{N}(CTeA_j),$\;
    $CTrN_j \gets TrN_j + (TeN_j - CTeN_j) + CTrA_j$\;

    \textbf{Step 2: Split Training Set} \\
    Divide $(CTrN_j, CTrA_j)$ into $k$ slices: $S_{j1}, S_{j2}, \dots, S_{jk}$

    \textbf{Step 3: Initialize and Train Model} \\
    $\theta_j \gets \theta_{\text{pretrained}}$\;

    \For{$i = 1$ \textbf{to} $k$}{
        Load and train on $S_{ji}$, updating $\theta_j$\;
    }

}

\Return $\theta_1, \theta_2, \dots, \theta_n$\;

\end{algorithm}

To align with the conventional anomaly detection training paradigm, we introduce HuVAD-S, constructed similarly to existing datasets \cite{liu2018future,rodrigues2020multi,danesh2023chad,Cao_2023_CVPR}. This subset serves as a foundation for the primary goal of this paper: advancing continual learning. This approach not only enables comparative analysis with traditional datasets but also allows for the application of the proposed Unsupervised Continual Anomaly Learning (UCAL) method to previous datasets, thereby broadening the impact of this new paradigm.


Following the principles of unsupervised anomaly detection, the training set is curated to contain only normal behavior. The test set, by contrast, includes both normal and anomalous behaviors in a 70/30 ratio to facilitate robust evaluation across varied conditions. As outlined in \cref{tab:frame-count}, HuVAD provides over five times more normal training frames than the next largest dataset, alongside a comprehensive test set exceeding 700K frames (approximately 6.5 hours), offering four times the scale of comparable datasets.

\begin{figure*}
    \centering
       \includegraphics[width=1\linewidth, trim= 28 80 38 40,clip]{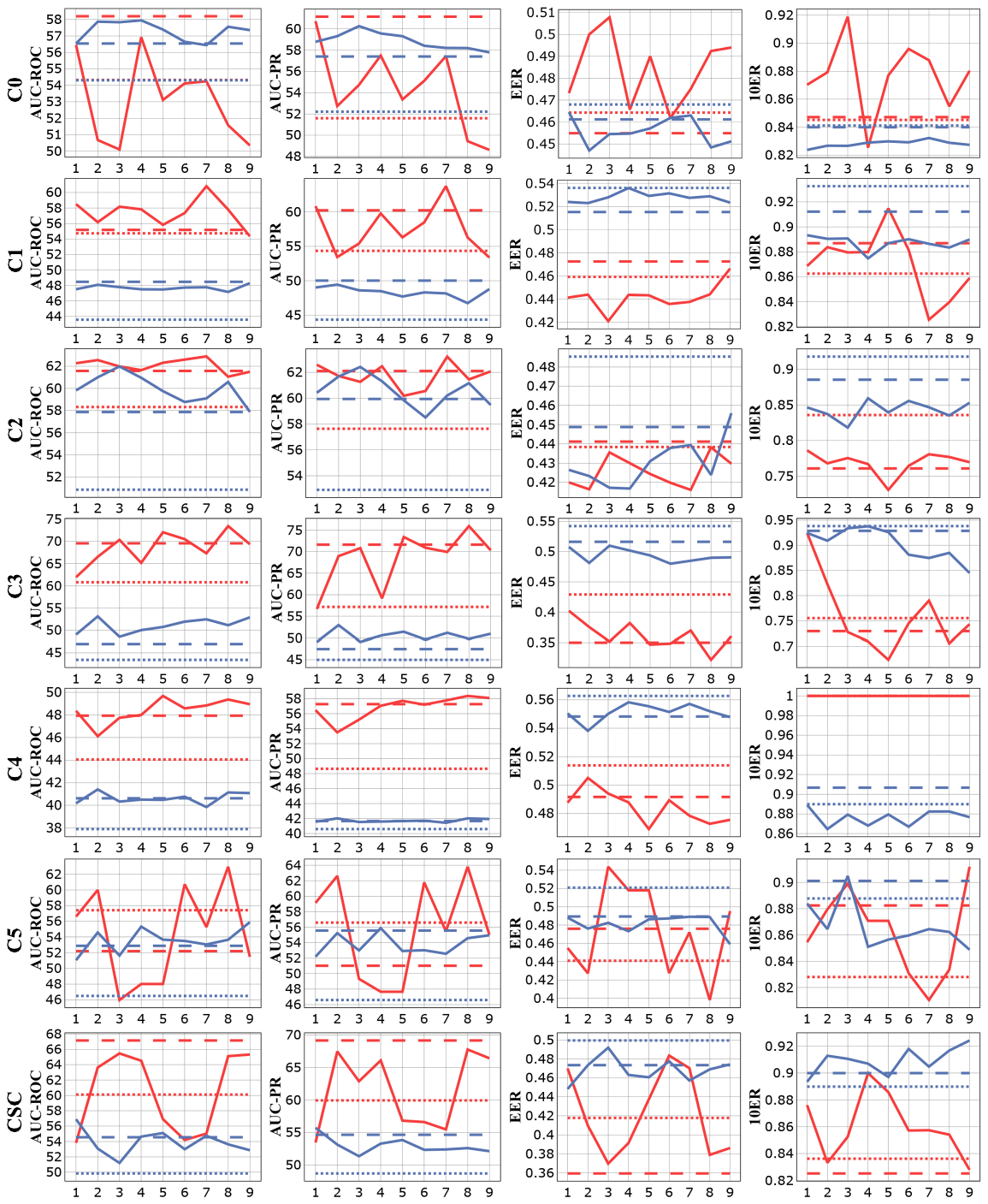}
            \caption{
                Continual learning benchmark using TSGAD (red) \cite{noghre2024exploratory} and STG-NF (blue) \cite{Hirschorn_2023_ICCV}. The dotted line is the baseline (trained on SHT's training set \cite{liu2018future} and tested on HuVAD-C's test set), the dashed line is standard training, and the solid line is continual learning.}
            
 \label{fig:big}
\end{figure*}

\subsection{Standard Results}
To demonstrate the effectiveness of the proposed dataset, we leverage SOTA pose-based anomaly detection models with an available code repository. Specifically, we employ MPED-RNN \cite{morais2019learning}, GEPC \cite{markovitz2020graph}, STG-NF \cite{Hirschorn_2023_ICCV} and TSGAD \cite{noghre2024exploratory} as discussed in \cref{sec:related}. For the TSGAD model, we opted to use only the pose branch, aligning with this study's primary focus on pose-based anomaly detection. As for all chosen models, hyper-parameters follow the original study, and detailed training parameters are available in the supplementary materials for replication and further investigation.



\begin{table*}[]
\small
\centering
\begin{tabular}{c|cccc|cccc}
\hline \hline
\textbf{Model} & \textbf{AUC-ROC} & \textbf{AUC-PR} & \textbf{EER} & \multicolumn{1}{c|}{\textbf{10ER}} & \textbf{AUC-ROC} & \textbf{AUC-PR} & \textbf{EER} & \textbf{10ER} \\ \hline
\textbf{} & \multicolumn{4}{c|}{\textbf{C0}} & \multicolumn{4}{c}{\textbf{C1}} \\ \hline
\textbf{TSGAD} & \textbf{58.19} & \textbf{61.14} & \textbf{0.45} & 0.84 & 55.17 & 60.25 & 0.47 & 0.88 \\
\textbf{ TSGAD+UCAL(Ours)} & 56.45 & 60.72 & 0.46 & \textbf{0.82} & \textbf{60.81} & \textbf{63.77} & \textbf{0.42} & \textbf{0.82} \\ \hline
\textbf{STG-NF} & 56.54 & 57.40 & 0.46 & 0.84 & \textbf{48.47} & \textbf{50.01} & \textbf{0.51} & 0.91 \\
\textbf{STG-NF+UCAL(Ours)} & \textbf{57.94} & \textbf{60.25} & \textbf{0.44} & \textbf{0.82} & 48.30 & 49.42 & 0.52 & \textbf{0.87} \\ \hline
\textbf{} & \multicolumn{4}{c|}{\textbf{C2}} & \multicolumn{4}{c}{\textbf{C3}} \\ \hline
\textbf{TSGAD} & 61.57 & 62.10 & 0.44 & 0.76 & 69.57 & 71.62 & 0.35 & 0.72 \\
\textbf{TSGAD+UCAL(Ours)} & \textbf{62.88} & \textbf{63.22} & \textbf{0.41} & \textbf{0.73} & \textbf{73.47} & \textbf{75.93} & \textbf{0.32} & \textbf{0.70} \\ \hline
\textbf{ STG-NF} & 57.86 & 59.94 & 0.44 & 0.88 & 46.96 & 47.45 & 0.51 & 0.92 \\
\textbf{ STG-NF+UCAL(Ours)} & \textbf{61.97} & \textbf{62.42} & \textbf{0.41} & \textbf{0.81} & \textbf{53.11} & \textbf{52.93} & \textbf{0.48} & \textbf{0.84} \\ \hline
\textbf{} & \multicolumn{4}{c|}{\textbf{C4}} & \multicolumn{4}{c}{\textbf{C5}} \\ \hline
\textbf{TSGAD} & 47.93 & 57.27 & 0.49 & 1.00 & 52.19 & 51.01 & 0.47 & 0.88 \\
\textbf{TSGAD+UCAL(Ours)} & \textbf{49.68} & \textbf{58.37} & \textbf{0.46} & \textbf{1.00} & \textbf{62.95} & \textbf{63.81} & \textbf{0.39} & \textbf{0.81} \\ \hline
\textbf{ STG-NF} & 40.62 & 41.67 & 0.54 & 0.90 & 52.87 & \textbf{55.56} & 0.48 & 0.90 \\
\textbf{ STG-NF+UCAL(Ours)} & \textbf{41.40} & \textbf{42.03} & \textbf{0.53} & \textbf{0.80} & \textbf{55.89} & 55.23 & \textbf{0.47} & \textbf{0.84} \\ \hline
\textbf{} & \multicolumn{8}{c}{\textbf{CSC}} \\ \hline
{} & \multicolumn{2}{c}{\textbf{AUC-ROC}} & \multicolumn{2}{c}{\textbf{AUC-PR}} & \multicolumn{2}{c}{\textbf{EER}} & \multicolumn{2}{c}{\textbf{10ER}} \\ \hline
\textbf{TSGAD} & \multicolumn{2}{c}{\textbf{67.15}} & \multicolumn{2}{c}{\textbf{69.14}} & \multicolumn{2}{c}{\textbf{0.35}} & \multicolumn{2}{c}{\textbf{0.82}} \\
\textbf{ TSGAD+UCAL(Ours)} & \multicolumn{2}{c}{65.46} & \multicolumn{2}{c}{67.75} & \multicolumn{2}{c}{0.36} & \multicolumn{2}{c}{\textbf{0.82}} \\ \hline
\textbf{ STG-NF} & \multicolumn{2}{c}{54.55} & \multicolumn{2}{c}{54.63} & \multicolumn{2}{c}{0.47} & \multicolumn{2}{c}{0.90} \\
\textbf{ STG-NF+UCAL(Ours)} & \multicolumn{2}{c}{\textbf{56.90}} & \multicolumn{2}{c}{\textbf{55.71}} & \multicolumn{2}{c}{\textbf{0.44}} & \multicolumn{2}{c}{\textbf{0.89}} \\ \hline \hline
\end{tabular}%
\caption{Continual learning benchmarks using TSGAD \cite{noghre2024exploratory} and STG-NF \cite{Hirschorn_2023_ICCV}. Results for TSGAD and STG-NF are based on conventional training and evaluation on the HuVAD-C. TSGAD+UCAL(Ours) and STG-NF+UCAL(Ours) show the best results of UCAL.}
\label{tab:continual}
\end{table*}


\cref{tab:benchmark} illustrates a comprehensive analysis of models benchmarked on the HuVAD dataset. MPED-RNN \cite{morais2019learning} consistently achieves the highest overall performance, both across combined and individual cameras, with TSGAD \cite{noghre2024exploratory} consistently ranking second. STG-NF \cite{Hirschorn_2023_ICCV} shows irregular performance, with the highest EER, lowest AUC-ROC, and highest AUC-PR across all experiments. 

\cref{tab:benchmark} also reveals that the CSC camera offering a parking lot view (see \cref{fig:camview}) is the most challenging for algorithms. As noted in \cref{sec:data_collection}, this camera captures context-specific behaviors such as security training, making it harder to distinguish anomalies in the test set. \cref{fig:full_analyze} also reveals that CSC has the highest crowd density and occlusion, further complicating its challenge for VAD models. These characteristics further highlight the need for solutions such as continual anomaly learning.

While EER indicates the lowest equal FNR and FPR possible, real-world VAD often prioritizes minimizing FNR. Therefore, a model with a lower 10ER is preferred \cite{ardabili2024exploring}. \cref{tab:benchmark} shows that a lower EER does not always align with a lower 10ER. Comparisons on C4 and CSC highlight that models with similar EERs can have different 10ERs, underscoring the importance of 10ER in model selection.


\section{Unsupervised Continual Anomaly Learning}

\label{sec:continual}
As discussed in \cref{sec:intro}, the context-specific and open-set nature of unsupervised anomaly detection necessitates learning of the normal distribution within each unique environment, limiting model generalizability. Consequently, models trained on specific datasets often lose effectiveness in real-world applications. To address this, we propose an Unsupervised Continual Anomaly Learning framework for each camera available in HuVAD that incrementally evolves the anomaly model, offering a more adaptive solution.

\subsection{Methodology}
\cref{fig:cal} illustrates the UCAL framework, which configures HuVAD-S to simulate real-world streaming data and performs continual anomaly learning. It begins by separating data by camera, with the Continual Data Rearrangement module applied independently to each camera’s dataset. Since anomalies are rare in the real world, Continual Data Rearrangement starts with HuVAD-S (\cref{sec:benchmark}) and injects anomalies randomly into the training stream, maintaining an anomaly ratio below 1\%, as specified in \cref{algo:continual_learning}. The test set is balanced to achieve a 1:1 ratio of normal to anomalous frames, enhancing metrics such as AUC-ROC and AUC-PR. Normal frames removed from the test set are added to the training stream as detailed in \cref{algo:continual_learning}. The new training and test sets form the HuVAD Continual benchmark (HuVAD-C). More details on the test and training sets for each camera are in the supplementary material.

The HuVAD-C training set is then divided into $k$ slices for $k$ training steps. As depicted in \cref{algo:continual_learning}, the UCAL training process begins with a pre-trained model from an external dataset and incrementally trains on the HuVAD-C training set. Model performance is tested on the HuVAD-C test set at each step. This structured, incremental approach allows for gradual model adaptation, effectively addressing the complexities of continual learning in VAD (see \cref{fig:cal}).



\subsection{UCAL Results}

We evaluate the two most recent VAD models, STG-NF \cite{Hirschorn_2023_ICCV} and TSGAD (pose branch) \cite{noghre2024exploratory}, using this framework. To simulate distribution shifts, we pre-train models on the SHT dataset \cite{liu2018future} as the origin data domain. For our experiments, we chose the number of slices (k) equal to 9 and trained the models for 9 steps with the learning rate of $5e-3$. More detailed parameters are provided in the supplementary material to ensure reproducibility.

The results of continual learning benchmarking are illustrated in \cref{fig:big}. The baseline, represented by the dotted lines, involves models trained on SHT \cite{liu2018future} and tested on the HuVAD-C test set without any fine-tuning for HuVAD-C. The dashed lines indicate conventional training and testing on HuVAD-C train and test sets. Our findings reveal that continual training outperforms the baseline in 98.21\% of cases, demonstrating significant performance improvement. Furthermore, as depicted in \cref{tab:continual} a comparison with conventional training shows that UCAL yields better results in 82.14\% of cases, highlighting its advantages over standard training. The HuVAD dataset has enabled benchmarking of continual learning, further validating these improvements.
\section{Conclusion}
\label{sec:conclusion}

This study presents HuVAD and UCAL, introducing a novel dataset and continual learning framework as a new paradigm for enhancing the effectiveness of VAD. By enabling continuous model adaptation, this paper addresses the limitations of conventional static benchmarks and paves the way for VAD systems that remain robust in dynamic, real-world environments. HuVAD also prioritizes privacy with comprehensive annotations across varied scenes. Our evaluation with both standard and continual benchmark, highlights the potential to drive future research in anomaly detection through adaptive, context-aware approaches.

\section*{Acknowledgement}
This research is funded and supported by the United States National Science Foundation (NSF) under award numbers 1831795 and 2329816

\bibliographystyle{IEEEtran} 
\bibliography{main} 

\end{document}